\documentclass[10pt,twocolumn,letterpaper]{article}

\usepackage{cvpr}
\usepackage{times}
\usepackage{epsfig}
\usepackage{amsmath}
\usepackage{amssymb}
\usepackage{amsthm}
\usepackage{amsfonts}       
\usepackage{bm}

\usepackage{graphicx}
\usepackage{caption}
\usepackage[labelformat=simple]{subcaption}
\usepackage{multirow}

\usepackage{booktabs}       
\usepackage[flushleft]{threeparttable}
\usepackage{authblk}

\usepackage[pagebackref=true,breaklinks=true,letterpaper=true,colorlinks,bookmarks=false]{hyperref}
\hypersetup{
   colorlinks=true,
   linkcolor=blue,
   citecolor=blue,
   filecolor=blue,
   urlcolor=blue}
\usepackage[nameinlink, capitalise]{cleveref}
\crefformat{section}{#2section~#1#3}
\Crefformat{section}{#2Section~#1#3}

\theoremstyle{definition}
\newtheorem{definition}{Definition}
\cvprfinalcopy 

\def\cvprPaperID{4744} 

\ifcvprfinal\fi
\begin{document}

\title{Label Efficient Semi-Supervised Learning via Graph Filtering}

\author[1]{Qimai Li}
\author[1]{Xiao-Ming Wu\thanks{Corresponding author}}
\author[1]{Han Liu}
\author[1]{Xiaotong Zhang}
\author[12]{Zhichao Guan}

\affil[1]{The Hong Kong Polytechnic University\\ $^2$Zhejiang University\\}
\affil[ ]{\tt\small \{csqmli,csxmwu,cshliu,csxtzhang\}@comp.polyu.edu.hk, zcguan@zju.edu.cn}

\maketitle

\begin{abstract}
Graph-based methods have been demonstrated as one of the most effective approaches for semi-supervised learning, as they can exploit the connectivity patterns between labeled and unlabeled data samples to improve learning performance. However, existing graph-based methods either are limited in their ability to jointly model graph structures and data features, such as the classical label propagation methods, or require a considerable amount of labeled data for training and validation due to high model complexity, such as the recent neural-network-based methods. In this paper, we address label efficient semi-supervised learning from a graph filtering perspective. Specifically, we propose a graph filtering framework that injects graph similarity into data features by taking them as signals on the graph and applying a low-pass graph filter to extract useful data representations for classification, where label efficiency can be achieved by conveniently adjusting the strength of the graph filter. Interestingly, this framework unifies two seemingly very different methods -- label propagation and graph convolutional networks. Revisiting them under the graph filtering framework leads to new insights that improve their modeling capabilities and reduce model complexity. Experiments on various semi-supervised classification tasks on four citation networks and one knowledge graph and one semi-supervised regression task for zero-shot image recognition validate our findings and proposals.
\end{abstract}


\section{Introduction}\label{sec:intro}

The success of deep learning and neural networks comes at the cost of large amounts of labeled data and long training time. Semi-supervised learning \cite{chapelle2006semi} is important as it can leverage ample available unlabeled data to aid supervised learning, thus greatly saving the cost, trouble, and time for human labeling. Many researches have shown that when used properly, unlabeled data can significantly improve learning performance \cite{zhu2009introduction,kingma2014semi,kipf2016semi}.

One effective approach to making use of unlabeled data is to represent the data in a graph where each labeled or unlabeled sample is a vertex and then model the relations between vertices. For some applications such as social network analysis, data exhibits a natural graph structure. For some other applications such as image or text classification, data may come in a vector form, and a graph is usually constructed using data features. Nevertheless, in many cases, graphs only partially encode data information. Take document classification in a citation network as an example, the citation links between documents form a graph which represents their citation relation, and each document is represented as a bag-of-words feature vector which describes its content. To correctly classify a document, both the citation relations and the content information need to be taken into account, as they contain different aspects of document information. For graph-based semi-supervised learning, the key challenge is to exploit graph structures as well as other information especially data features to improve learning performance.

Despite many progresses, existing methods are still limited in their capabilities to leverage multiple modalities of data information for learning. The classical label propagation methods only exploit graph structures to make predictions on unlabeled examples, which are often inadequate in many scenarios. To go beyond their limit and jointly model graph structures and data features, a common approach is to train a supervised learner to classify data features while regularizing the classifier with graph similarity. Manifold regularization  \cite{belkin2006manifold} trains a support vector machine with a graph Laplacian regularizer. Deep semi-supervised embedding \cite{weston2012deep} and Planetoid \cite{yang2016revisiting} train a neural network with an embedding-based regularizer. Recently, graph convolutional networks (GCN) \cite{kipf2016semi} have demonstrated impressive results in semi-supervised learning, due to its special design of a first-order convolutional filter that nicely integrates graph and feature information in each layer. The success of GCN has inspired many follow-up works \cite{corr/abs-1812-04202,corr/abs-1812-08434} on graph neural networks for semi-supervised learning. However, although these neural-network-based models tend to have stronger modelling capabilities than the conventional ones, they typically require a considerable amount of labeled data for training and validation due to high model complexity, hence may not be label efficient.

In this paper, we propose to study semi-supervised learning from a principled graph filtering perspective. The basic idea is to take data features as signals sitting on the underlying graph that encodes relations between data samples, and uses the graph to design proper low-pass graph convolutional filters to generate smooth and representative features for subsequent classification. In this process, graph similarity is injected into data features to produce more faithful data representations. It also enables learning with few labels by flexibly adjusting filter strength to achieve label efficiency. More interestingly, it unifies well-known semi-supervised learning methods including the label propagation methods \cite{Zhou03} and the graph convolutional networks \cite{kipf2016semi}, with useful insights for improving their modelling capabilities.

Under the graph filtering framework, we show that label propagation methods can be decomposed into three components: graph signal, graph filter, and classifier. Based on this, we then propose a class of generalized label propagation (GLP) methods by naturally extending the three components, including using data feature matrix instead of label matrix as input graph signals, extending the graph filter to be any low-pass graph convolutional filter, and using any desired classifier for classification. GLP can achieve label efficiency in semi-supervised learning by taking advantages of data features, strong and efficient graph filters, and powerful supervised classifiers.

The popular graph convolutional networks (GCN) \cite{kipf2016semi} can also be interpreted under the graph filtering framework. It has been shown that GCN implements the graph convolution in each layer by conducting Laplacian smoothing \cite{li2018deeper}. When revisited under the graph filtering framework, it further elucidates the inner working of GCN including the renormalization trick and the model parameter settings. Furthermore, it leads to an improved GCN model (IGCN) that is more flexible and label efficient. By adding an exponent parameter on the filter of GCN to easily control filter strength, IGCN can significantly reduce trainable parameters and effectively prevent overfitting when training data is very limited.

We conduct extensive experiments to validate our findings and the effectiveness of the proposed methods. We test a variety of semi-supervised classification tasks including document classification on four citation networks and entity classification on one knowledge graph. We also test a semi-supervised regression task for zero-shot image recognition. The proposed GLP and IGCN methods perform superiorly in terms of prediction accuracy and training efficiency.

Our contributions are summarized as follows. First, we propose a graph filtering framework for semi-supervised learning, which provides new insights into GCN and shows its close connection with label propagation. Second, we propose GLP and IGCN to successfully tackle label insufficiency in semi-supervised learning. Third, we demonstrate the high efficacy and efficiency of the proposed methods on various semi-supervised classification and regression tasks.

\section{Graph Filtering}\label{sec:GSP}

This section introduces the concepts of graph signal, graph convolutional filter, and graph filtering.

\textbf{Notations.} A non-oriented graph $\mathcal{G}=(\mathcal{V},W, X)$ with $n=|\mathcal{V}|$ vertices is given, with a nonnegative symmetric affinity matrix $W\in \mathbb{R}^{n\times n}_+$ encoding edge weights and a feature matrix $X\in \mathbb{R}^{n\times m}$ where an $m$-dimensional feature vector is associated with each vertex. For semi-supervised classification, only a small subset of vertices are labeled, and the goal is to predict the labels of other vertices. Denote by $Y\in \{0,1\}^{n\times l}$ the label matrix\footnote{If the label of vertex $v_i$ is known, then $Y(i,:)$ is a one-hot embedding of $v_i$ with $y_{ij}=1$ if $v_i$ belongs to the $j$-th class and $y_{ij}=0$ otherwise. If the label of vertex $v_i$ is not given, then $Y(i,:)$ is a vector of all zeros.}, where $l$ is the number of classes.

In graph signal processing \cite{shuman2013emerging}, the eigenvalues and eigenvectors of the graph Laplacian correspond to frequencies and Fourier basis in classical harmonic analysis. The graph Laplacian is defined as $L=D-W$, where $D$ is the degree matrix. It can be eigen-decomposed as $L=\Phi\Lambda\Phi^{-1}$, where $\Lambda=\text{diag}(\lambda_1,\cdots,\lambda_n)$ and $(\lambda_i)_{1\le i \le n}$ are the eigenvalues in the increasing order, and $\Phi=(\bm\phi_1,\cdots,\bm\phi_n)$ and $(\bm\phi_i)_{1\le i \le n}$ are the associated orthogonal eigenvectors. Note that the normalized graph Laplacian $L_{r}=D^{-1}L$ and the symmetrically normalized graph Laplacian $L_{s}=D^{-\frac12}LD^{-\frac12}$ have similar eigen-decomposition as $L$. The eigenvalues $(\lambda_i)_{1\le i \le n}$ can be considered as frequencies, and the associated eigenvectors $ (\bm\phi_i)_{1\le i \le n}$ form the Fourier basis.
\begin{definition}[Graph Signal]
A \emph{graph signal} is a real-valued function $f: \mathcal{V} \to \mathbb{R}$ on the vertex set $\mathcal{V}$ of a graph, which can be represented as $\bm f=(f(v_1),\cdots,f(v_n))^\top$ in the vector form.
\end{definition}

Any graph signal $\bm f$ can be decomposed into a linear combination of the basis signals $\bm (\phi_i)_{1\le i \le n}$ :
\begin{equation}\label{eq:signal-decomp}
    \bm f=\Phi \bm c =\sum_i{c_i \bm \phi_i},
\end{equation}
where $\bm c = (c_1,\cdots, c_n)^\top$ and $c_i$ is the coefficient of $\bm \phi_i$. The magnitude of the coefficient $|c_i|$ represents the strength of the basis signal $\bm \phi_i$ presented in the signal $\bm f$. It is well known that the basis signals associated with lower frequencies (smaller eigenvalues) are smoother \cite{zhu2009introduction} on the graph, where the smoothness of the basis signal $\bm \phi_i$ is measured by the eigenvalue $\lambda_i$, i.e.,
\begin{equation} \label{eq:laplacian_property}
\sum_{(v_j,v_k)\in \mathcal{E}}{w_{jk}[\bm \phi_{i}(j)-\bm \phi_{i}(k)]^2} = {\bm \phi_i}^\top L{\bm \phi_i} =\lambda_i.
\end{equation}
Hence, a smooth graph signal $\bm f$ should mostly consist of low-frequency basis signals.

The basic idea of graph filtering is to use the underlying data relation graph to design proper graph filters to produce smooth signals for downstream tasks. A graph filter is a function that takes a graph signal as input and outputs a new signal. A linear graph filter can be represented as a matrix $G\in \mathbb{R}^{n\times n}$, and the output signal is $G\bm f$. In this paper, we focus on graph convolutional filters due to their linear shift-invariant property \cite{sandryhaila2013discrete}.

\begin{definition}[Graph Convolutional Filter] \label{eq:Graph Convolutional Filter}
    A linear graph filter $G$ is convolutional if and only if there exists a function $p(\cdot):\mathbb R \to \mathbb R$, satisfying $G=\Phi p(\Lambda)\Phi^{-1}$, where $p(\Lambda)=\text{diag}(p(\lambda_1),\cdots,p(\lambda_n))$.
\end{definition}
The function $p(\cdot)$ is known as the \emph{frequency response function} of the filter $G$. We shall denote by $p(L)$ the filter with frequency response function $p(\cdot)$.

\begin{figure}[t]
	\centering
	\begin{subfigure}{0.23\textwidth}
		\includegraphics[width=\textwidth]{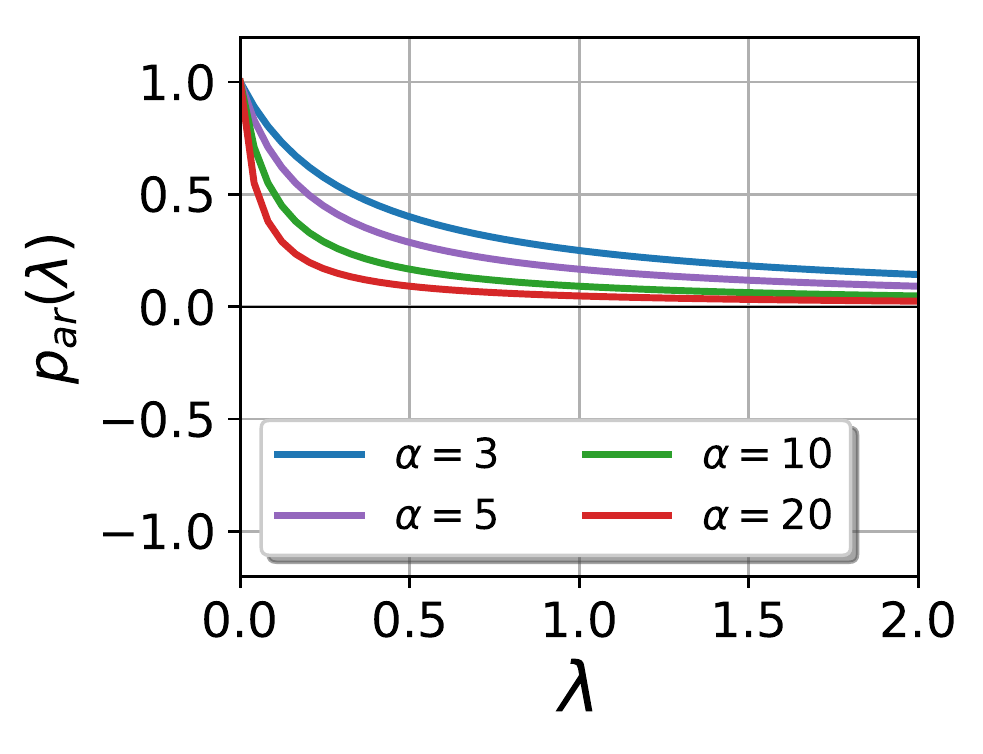}
		\caption{$p_{\text{ar}}(\lambda)=(1+\alpha\lambda)^{-1}$}
		\label{fig:AP}
	\end{subfigure}
	\begin{subfigure}{0.23\textwidth}
		\includegraphics[width=\textwidth]{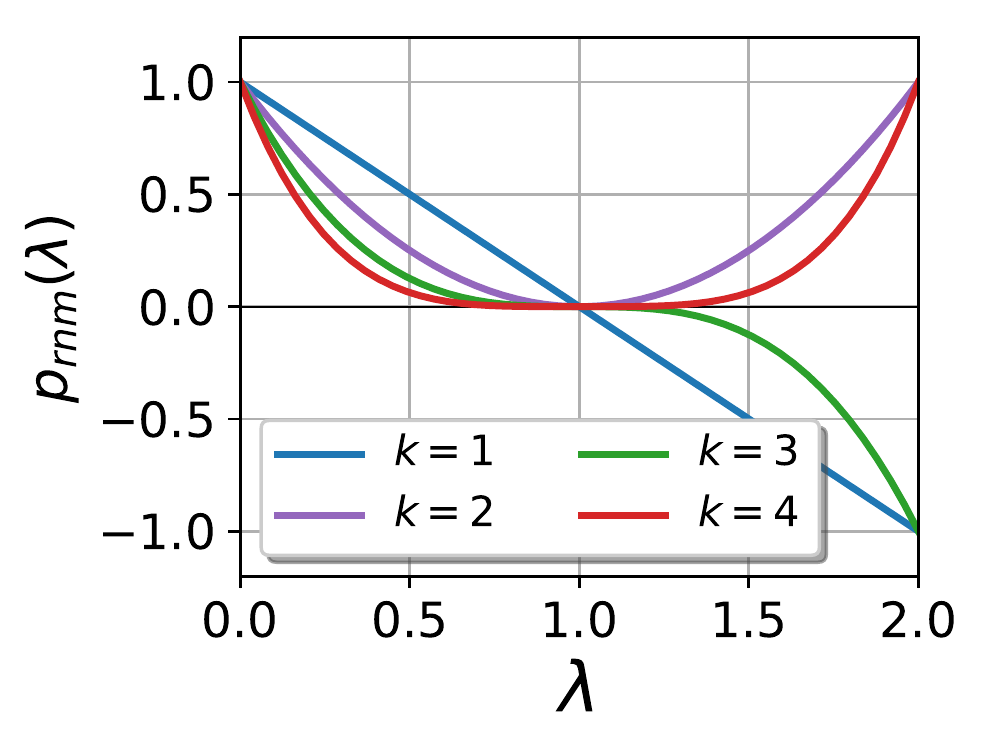}
		\caption{$p_\text{rnm}(\lambda)=(1-\lambda)^k$}
		\label{fig:RNM}
	\end{subfigure}
	\caption{Frequency response functions.}
	\label{fig:filters}
\end{figure}

To produce a smooth signal, the filter $G$ should be able to preserve the low-frequency basis signals in $\bm f$ while filtering out the high-frequency ones. By \eqref{eq:signal-decomp}, the output signal can be written as
\begin{equation}
\bar{\bm f}= G\bm f =\Phi p(\Lambda) \Phi^{-1} \cdot \Phi \bm c=\sum_i{p(\lambda_i)c_i \bm \phi_i}. \label{eq:filter_decom}
\end{equation}
In the output signal $\bar{\bm f}$, the coefficient $c_i$ of the basis signal $\bm \phi_i$ is scaled by $p(\lambda_i)$. To preserve the low-frequency signals and remove the high-frequency ones, $p(\lambda_i)$ should be large for small $\lambda_i$ and small for large $\lambda_i$. Simply put, $p(\cdot)$ should behave like a \emph{low-pass} filter in classical harmonic analysis. \cref{fig:AP} shows an example of a low-pass  function whose response decreases as the frequency increases.

Taking the vertex features as graph signals, e.g., a column of the feature matrix $X$ can be considered as a graph signal, graph filtering provides a principled way to integrate graph structures and vertex features for learning. In the following, we will revisit two popular semi-supervised learning methods -- label propagation and graph convolutional networks under this framework and gain new insights for improving their modelling capabilities.

\section{Revisit and Extend Label Propagation}\label{sec:GLP}

Label propagation (LP) \cite{Zhu03,Zhou03,bengio2006label} is arguably the most popular method for graph-based semi-supervised learning. As a simple and effective tool, it has been widely used in many scientific research fields and numerous industrial applications. The objective of LP is to find a prediction (embedding) matrix $Z\in \mathbb{R}^{n\times l}$ that agrees with the label matrix $Y$ while being smooth on the graph such that nearby vertices have similar embeddings:
\begin{equation}\label{eq:lp optimization}
Z=\operatorname*{arg\,min}_{Z}\{\underbrace{||Z-Y||_2^2}_{\text{Least square fitting}}\;+\; \underbrace{\alpha\text{Tr}(Z^\top L Z)}_{\text{Laplcacian regularization}}\},
\end{equation}
where $\alpha$ is a balancing parameter controlling the degree of Laplacian regularization. In \eqref{eq:lp optimization}, the fitting term enforces the prediction matrix $Z$ to agree with the label matrix $Y$, while the regularization term makes each column of $Z$ smooth along the graph edges. A closed-form solution of the above unconstrained quadratic optimization can be obtained by taking the derivative of the objective function and setting it to zero:
\begin{equation}\label{eq:lp solution}
    Z=(I+\alpha L)^{-1}Y.
\end{equation}
Each unlabeled vertex $v_i$ is then classified by simply comparing the elements in $Z(i,:)$ or with some normalization applied on the columns of $Z$ first \cite{Zhu03}.

\subsection{Revisit Label Propagation}

From the perspective of graph filtering, we show that LP is comprised of three components: signal, filter, and classifier. We can see from \eqref{eq:lp solution} that the input signal matrix of LP is simply the label matrix $Y$, where each column $Y(:,i)$ can be considered as a graph \emph{signal}. Note that in $Y(:,i)$, only the labeled vertices in class $i$ have value 1 and others 0.

The graph \emph{filter} of LP is the Auto-Regressive (AR) filter \cite{tremblay2018design}:
\begin{equation}\label{eq:ar}
	p_{\text{ar}}(L)=(I+\alpha L)^{-1}
	=\Phi (I+\alpha \Lambda)^{-1}\Phi^{-1},
\end{equation}
with the frequency response function:
\begin{equation}\label{eq:lp filter}
p_{\text{ar}}(\lambda_i)=\frac{1}{1+\alpha\lambda_i}.
\end{equation}
Note that this also holds for the normalized graph Laplacians. As shown in \cref{fig:AP}, $p_{\text{ar}}(\lambda_i)$ is low-pass. For any $\alpha >0$, $p_{\text{ar}}(\lambda_i)$ is near 1 when $\lambda_i$ is close to 0, and $p_{\text{ar}}(\lambda_i)$ decreases and approaches 0 as $\lambda_i$ increases. Applying the AR filter on the signal $Y(:,i)$ will produce a smooth signal $Z(:,i)$, where vertices of the same class have similar values and those of class $i$ have larger values than others under the cluster assumption. The parameter $\alpha$ controls the strength of the AR filter. When $\alpha$ increases, the filter becomes more low-pass (\cref{fig:AP}) and will produce smoother signals.

Finally, LP adopts a nonparametric \emph{classifier} on the embeddings to classify the unlabeled vertices, i.e., the label of an unlabeled vertex $v_i$ is given by $y_i=\arg\max_jZ(i,j)$.

\subsection{Generalized Label Propagation Methods}\label{sec:model}

The above analysis shows that LP only takes into account the given graph $W$ and the label matrix $Y$, but without using the feature matrix $X$. This is one of its major limitations in dealing with datasets that provide both $W$ and $X$, e.g., citation networks. Here, we propose generalized label propagation (GLP) methods by naturally extending the three components of LP.
\begin{itemize}
	\item Signal: Use the feature matrix $X$ instead of the label matrix $Y$ as input signals.
	\item Filter: The filter $G$ can be any low-pass graph convolutional filter.
	\item Classifier: The classifier can be any classifer trained on the embeddings of labeled vertices.
\end{itemize}
GLP consists of two simple steps. First, a low-pass filter $G$ is applied on the feature matrix $X$ to obtain a smooth feature matrix $\bar{X}\in \mathbb{R}^{n\times m}$:
\begin{equation}\label{eq:filter-G}
    \bar{X}=GX.
\end{equation}
Second, a supervised classifier (e.g., multilayer perceptron, convolutional neural networks, support vector machines, etc.) is trained with the filtered features of labeled vertices, which is then applied on the filtered features of unlabeled vertices to predict their labels.

GLP has the following advantages. First, by injecting graph relations into data features, it can produce more useful data representations for the downstream classification task. Second, it offers the flexibility of using computationally efficient filters and conveniently adjusting their strength for different application scenarios. Third, it allows taking advantage of powerful domain-specific classifiers for high-dimensional data features, e.g., a multilayer perceptron for text data and a convolutional neural network for image data.

\section{Revisit and Improve Graph Convolutional Networks}\label{sec:gcn}

The recently proposed graph convolutional networks (GCN) \cite{kipf2016semi} has demonstrated superior performance in semi-supervised learning and attracted much attention. The GCN model consists of three steps. First, a so-called renormalization trick is applied on the adjacency matrix $W$ by adding an self-loop to each vertex, resulting in a new adjacency matrix $\tilde{W}=W+I$ with the degree matrix $\tilde{D}=D+I$, which is then symmetrically normalized as $\tilde{W}_{s}=\tilde{D}^{-\frac12}\tilde{W}\tilde{D}^{-\frac12}$.
Second, define the layerwise propagation rule:
\begin{equation}\label{eq:convolution}
H^{(t+1)}= \sigma \left(\tilde{W}_{s}H^{(t)}\Theta^{(t)}\right),
\end{equation}
where $H^{(t)}$ is the matrix of activations fed to the $t$-th layer and $H^{(0)} = X$, $\Theta^{(t)}$ is the trainable weight matrix in the layer, and $\sigma$ is the activation function, e.g., $\text{ReLU}(\cdot) = \max(0, \cdot)$. The graph convolution is defined as multiplying the input of each layer with the renormalized adjacency matrix $\tilde{W}_{s}$ from the left, i.e., $\tilde{W}_{s}H^{(t)}$. The convoluted features are then fed into a projection matrix $\Theta^{(t)}$. Third, stack two layers up and apply a softmax function on the output features to produce a prediction matrix:
\begin{equation}\label{eq:kipfGCN}
Z = \text{softmax}\left (
    \tilde{W}_{s}\; \text{ReLU}\left(
    \tilde{W}_{s}X\Theta^{(0)}
    \right)\Theta^{(1)}
\right),
\end{equation}
and then train the model with the cross-entropy loss on labeled samples.

\subsection{Revisit Graph Convolutional Networks}\label{sec:gcn fp}

In this section, we interpret GCN under the graph filtering framework and explain its implicit design features including the choice of the normalized graph Laplacian and the renormalization trick on the adjacency matrix.

GCN conducts graph filtering in each layer with the filter $\tilde{W}_{s}$ and the signal matrix $H^{(t)}$. We have $\tilde{W}_{s}=I-\tilde{L}_{s}$, where $\tilde{L}_{s}$ is the symmetrically normalized graph Laplacian of the graph $\tilde{W}$. Eigen-decompose $\tilde{L}_{s}$ as $\tilde{L}_{s}={\Phi}{\tilde\Lambda}{\Phi^{-1}}$, then the filter is
\begin{equation}\label{eq:GCN filter}
\tilde{W}_{s}=I-\tilde{L}_{s}={\Phi} (I-{\tilde\Lambda}){\Phi^{-1}},
\end{equation}
with frequency response function
\begin{equation}\label{eq:GCN response}
   p(\tilde{\lambda_i})=1-\tilde{\lambda_i}.
\end{equation}
Clearly, as shown in \cref{fig:RNM}, this function is linear and low-pass on [0, 1], but not on [1, 2].

It can be seen that by performing all the graph convolutions in \eqref{eq:kipfGCN} first, i.e., by exchanging the renormalized adjacency matrix $\tilde{W}_{s}$ in the second layer with the internal ReLU function, GCN is a special case of GLP, where the input signal matrix is $X$, the filter is $\tilde{W}_{s}^2$, and the classifier is a two-layer multi-layer perceptron (MLP). One can also see that GCN stacks two convolutional layers because $\tilde{W}_{s}^2$ is more low-pass than $\tilde{W}_{s}$, which can be seen from \cref{fig:RNM} that $(1-\lambda)^2$ is sort of more low-pass than $(1-\lambda)$ by suppressing the large eigenvalues harder .

\textbf{Why Use Normalized Graph Laplacian.} GCN uses the normalized Laplacian $L_{s}$ because the eigenvalues of $L_{s}$ fall into $[0, 2]$ \cite{Chung97}, while those of the unnormalized Laplacian $L$ are in $[0, +\infty]$. If using $L$, the frequency response in \eqref{eq:GCN response} will amplify eigenvalues in $[2, +\infty]$, which will introduce noise and undermine performance.

\begin{figure}[t]
	\centering
	\begin{subfigure}{0.20\textwidth}
		\includegraphics[width=\textwidth]{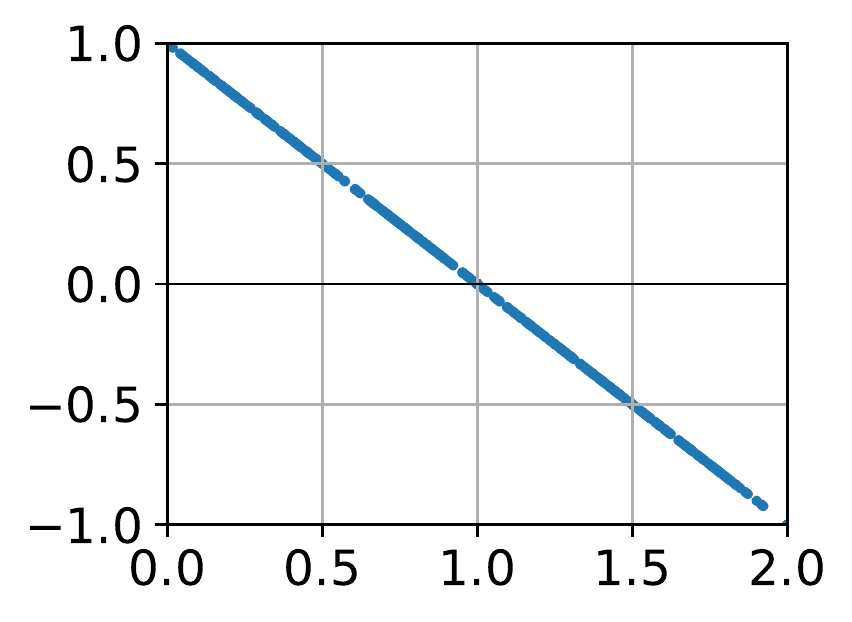}
		\caption{$1-\lambda$}
	\end{subfigure}
	\begin{subfigure}{0.20\textwidth}
		\includegraphics[width=\textwidth]{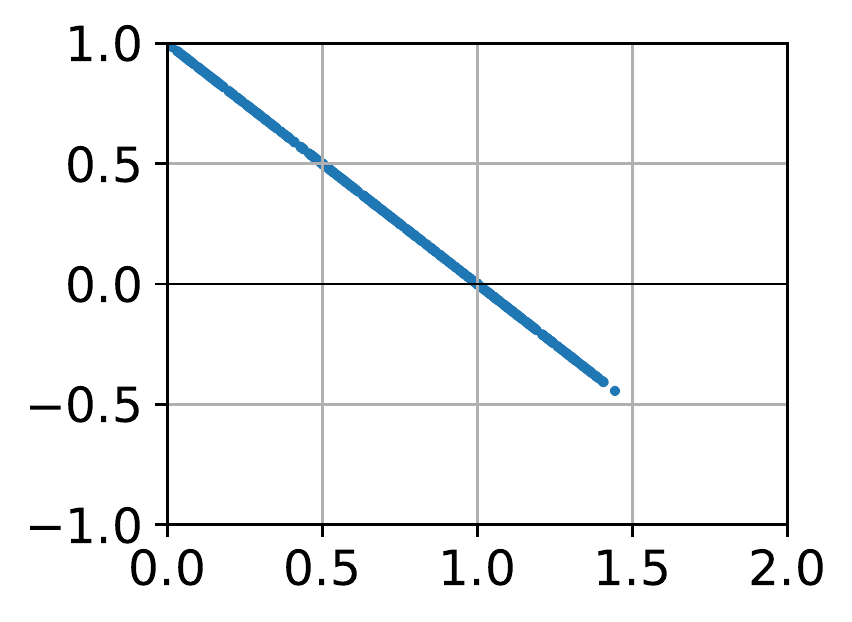}
		\caption{$1-\tilde{\lambda}$}
		\label{fig:renormalization}
	\end{subfigure}
	\begin{subfigure}{0.20\textwidth}
		\includegraphics[width=\textwidth]{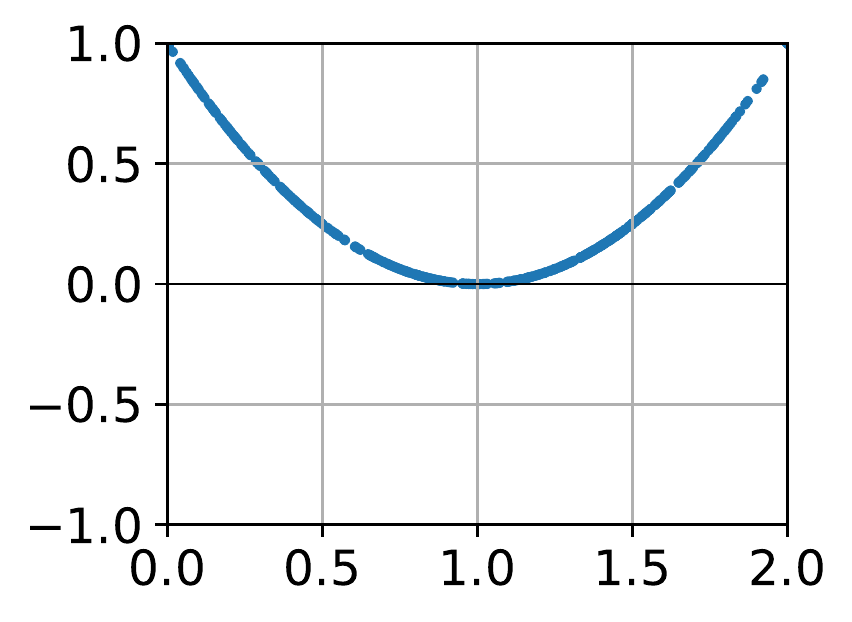}
		\caption{$(1-\lambda)^2$}
	\end{subfigure}
	\begin{subfigure}{0.20\textwidth}
		\includegraphics[width=\textwidth]{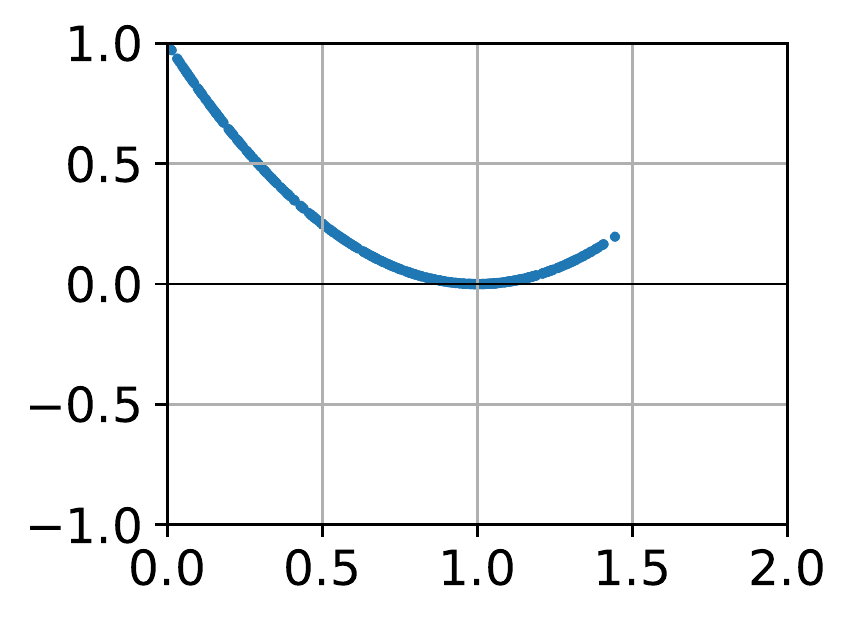}
		\caption{$(1-\tilde{\lambda})^2$}
	\end{subfigure}
	\caption{Effect of the renormalization trick. Left two figures plot points $(\lambda_i, p(\lambda_i))$. Right two figures plot points $(\tilde{\lambda}_i, p(\tilde{\lambda}_i))$.}
	\label{fig:compressing effect}
\end{figure}

\textbf{Why the Renormalization Trick Works.} We illustrate the effect of the renormalization trick used in GCN in \cref{fig:compressing effect}, where the frequency responses on the eigenvalues of $L_s$ and $\tilde{L_s}$ on the Cora citation network are plotted respectively. We can see that by adding a self-loop to each vertex, the range of eigenvalues shrinks from [0, 2] to [0, 1.5], which avoids amplifying eigenvalues near 2 and reduces noise. Hence, although the response function $(1-\lambda)^k$ is not completely low-pass, the renormalization trick shrinks the range of eigenvalues and makes $\tilde{L}_s$ resemble a low-pass filter. It can be proved that if the largest eigenvalue of $L_s$ is $\lambda_m$, then all the eigenvalues of $\tilde{L_s}$ are no larger than $\frac{d_m}{d_m+1}\lambda_m$, where $d_m$ is the largest degree of all vertices.

\subsection{Improved Graph Convolutional Networks}

One notable drawback of the current GCN model is that one cannot easily control filter strength. To increase filter strength and produce smoother features, one has to stack multiple layers. However, since in each layer the convolution is coupled with a projection matrix by the \text{ReLU}, stacking many layers will introduce many trainable parameters. This may lead to severe overfitting when label rate is small, or it will require extra labeled data for validation and model selection, both of which are not label efficient.

To fix this, we propose an improved GCN model (IGCN) by replacing the filter $\tilde{W}_{s}$ with $\tilde{W}_{s}^k$:
\begin{equation}\label{eq:kipfGCNk}
Z = \text{softmax}\left (
    \tilde{W}_{s}^k\; \text{ReLU}\left(
    \tilde{W}_{s}^kX\Theta^{(0)}
    \right)\Theta^{(1)}
\right).
\end{equation}
We call $p_\text{rnm}(\tilde{L}_{s})=\tilde{W}_{s}^k$ the renormalization (RNM) filter, with frequency response function
\begin{equation}\label{eq:filter gcn1}
p_{\text{rnm}}({\lambda})=\left(I-\tilde{\lambda}\right)^k.
\end{equation}
IGCN can achieve label efficiency by using the exponent $k$ to conveniently adjust the filter strength. In this way, it can maintain a shallow structure with a reasonable number of trainable parameters to avoid overfitting.

\section{Filter Strength and Computation}\label{sec:design}

The strength of the AR and RNM filters is controlled by the parameters $\alpha$ and $k$ respectively. However, choosing appropriate $\alpha$ and $k$ for different application scenarios is non-trivial. An important factor that should be taken into account is label rate. Intuitively, when there are very few labels in each class, one should increase filter strength such that distant nodes can have similar feature representations as the labeled nodes for the ease of classification. However, over-smoothing often results in inaccurate class boundaries. Therefore, when label rate is reasonably large, it would be desirable to reduce filter strength to preserve feature diversity in order to learn more accurate class boundaries.

\begin{figure}[t]
	\centering
	\begin{subfigure}{0.18\textwidth}
		\includegraphics[width=\textwidth]{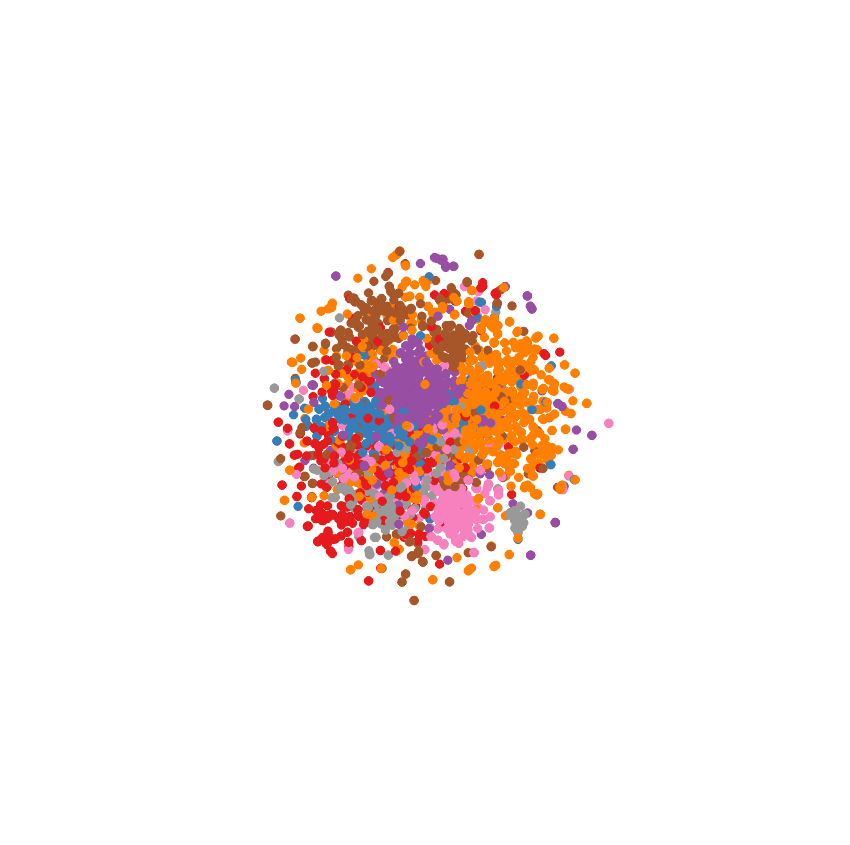}
		\caption{Raw features}
	\end{subfigure}
	\begin{subfigure}{0.18\textwidth}
		\includegraphics[width=\textwidth]{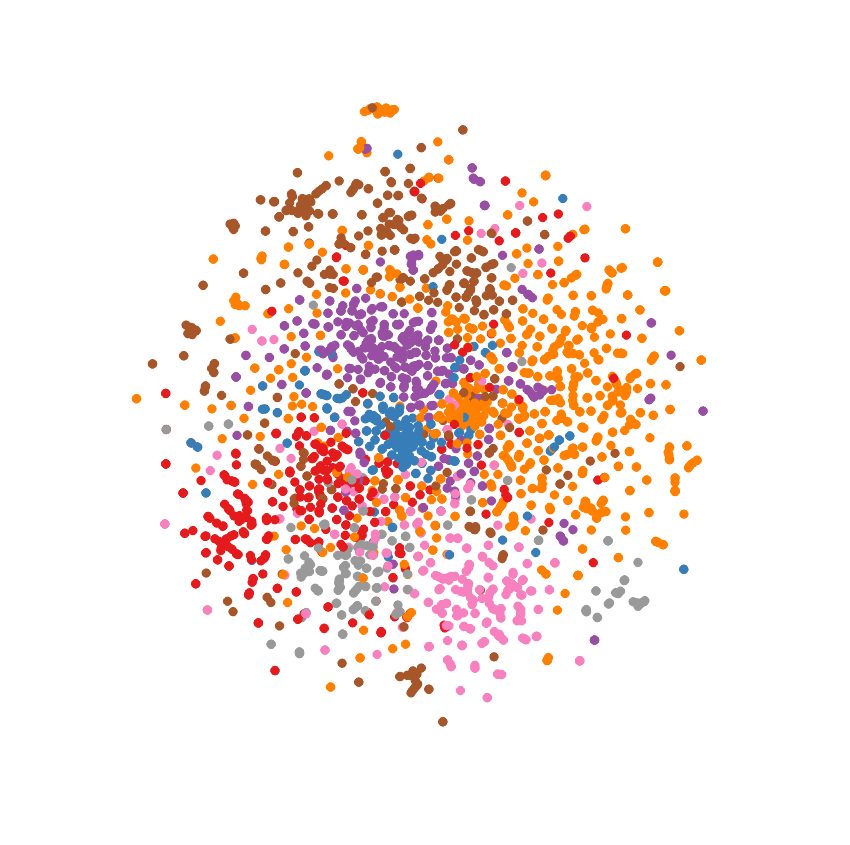}
		\caption{$k=1$}
	\end{subfigure}
	\begin{subfigure}{0.18\textwidth}
		\includegraphics[width=\textwidth]{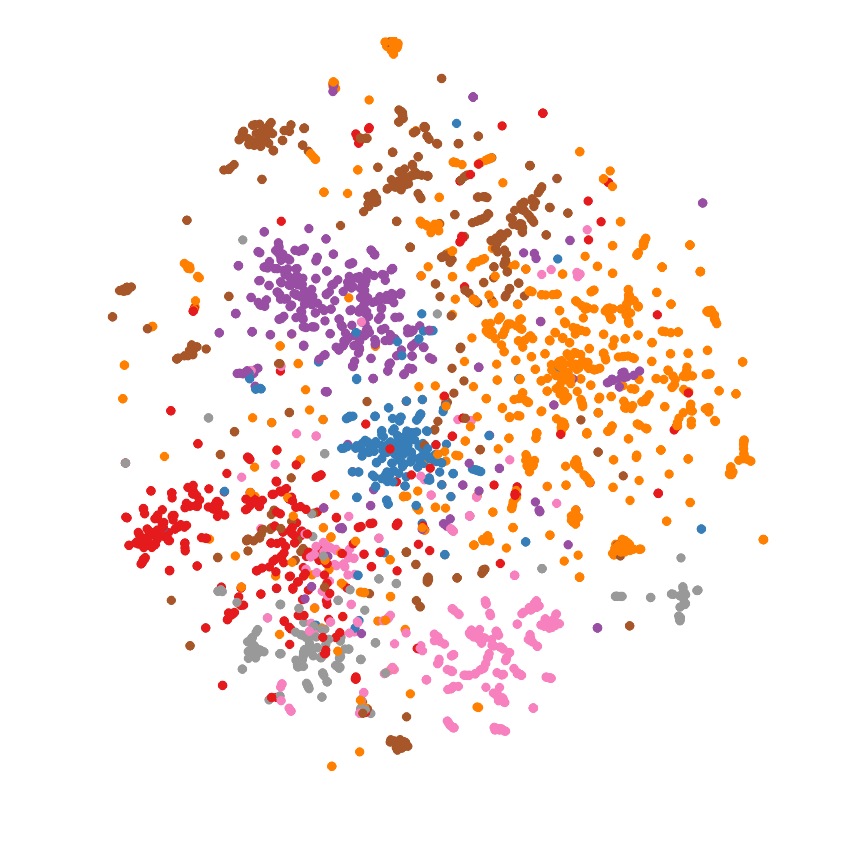}
		\caption{$k=5$}
	\end{subfigure}
	\begin{subfigure}{0.18\textwidth}
		\includegraphics[width=\textwidth]{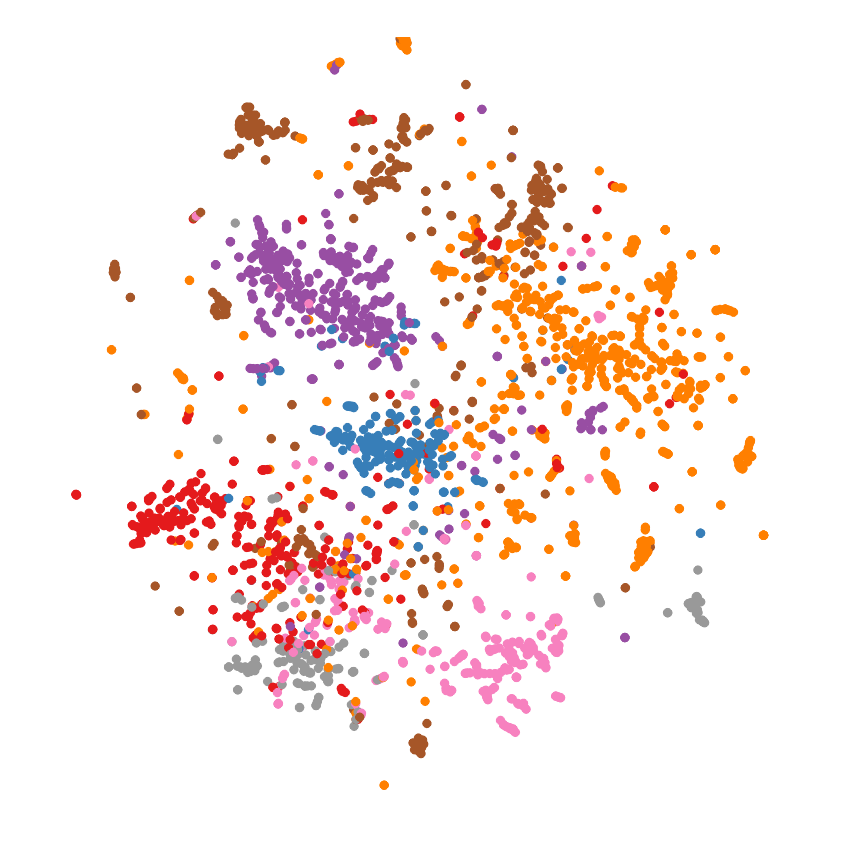}
		\caption{$k=10$}
	\end{subfigure}
	\caption{Visualization of raw and filtered Cora features (by using the RNM filter with different $k$).}
	\label{fig:cora_tsne}
\end{figure}

\cref{fig:cora_tsne} visualizes the raw and filtered features of Cora produced by the RNM filter and projected by t-SNE \cite{van2008visualizing}. It can be seen that as $k$ increases, the RNM filter produces smoother embeddings, i.e., the filtered features exhibit a more compact cluster structure, making it possible for classification with few labels.

The computation of the AR filter $p_{\text{ar}}(L)=(I+\alpha L)^{-1}$ involves matrix inversion, which is computationally expensive with complexity $\mathcal{O}(n^3)$. Fortunately, we can circumvent this problem by approximating $p_{\text{ar}}$ using its polynomial expansion:
\begin{equation}
    (I+\alpha L)^{-1}=\frac{1}{1+\alpha}\sum_{i=0}^{+\infty}{\left[\frac{\alpha}{1+\alpha}W\right]^i}, (\alpha>0).
\end{equation}
We can then compute $\bar{X}=p_{\text{ar}}(L)X$ iteratively with
$$X'^{(0)}={\bf O},\cdots,X'^{(i+1)}=X+\frac{\alpha}{1+\alpha}W X'^{(i)},$$
and let $\bar{X}=\frac{1}{1+\alpha}X'^{(k)}$. Empirically, we find that $k=\lceil4\alpha\rceil$ is enough to get a good approximation. Hence, the computational complexity is reduced to $\mathcal{O}(nm\alpha+Nm\alpha)$ (note that $X$ is of size $n\times m$), where $N$ is the number of nonzero entries in $L$, and $N\ll n^2$ when the graph is sparse.

For the RNM filter $p_{\text{rnm}}(\tilde{L}_s)=\tilde{W}_{s}^k=\left(I-\tilde{L}_s\right)^k$, note that for a sparse graph, $(I-\tilde{L}_s)$ is a sparse matrix. Hence, the fastest way to compute $\bar{X}=p_{\text{rnm}}(\tilde{L}_s)X$ is to left multiply $X$ by $(I-\tilde{L}_s)$ repeatedly for $k$ times, which has the computational complexity $\mathcal{O}(Nmk)$.

\section{Experiments}\label{sec:experiment}

To validate the performance of our methods GLP and IGCN, we conduct experiments on various semi-supervised classification tasks and a semi-supervised regression task for zero-shot image recognition.

\subsection{Semi-Supervised Classification}\label{sec:exp-citation}
For semi-supervised classification, we test our methods GLP and IGCN on two tasks.\footnote{Code is available at https://github.com/liqimai/Efficient-SSL} 1) Semi-supervised document classification on citation networks, where nodes are documents and edges are citation links. The goal is to classify the type of the documents with only a few labeled documents. 2) Semi-supervised entity classification on a knowledge graph. A bipartite graph is extracted from the knowledge graph \cite{yang2016revisiting}, and there are two kinds of nodes: entity and relation, where the edges are between the entity and relation nodes. The goal is to classify the entity nodes with only a few labeled entity nodes.

\noindent \textbf{Datasets.} We evaluate our methods on four citation networks -- Cora, CiteSeer, PubMed \cite{sen2008collective} and Large Cora, and one knowledge graph -- NELL \cite{carlson2010toward}. The dataset statistics are summarized in \cref{tab:app_dataset}. On citation networks, we test two scenarios -- 4 labels per class and 20 labels per class. On NELL, we test three scenarios -- 0.1\%, 1\% and 10\% label rates.

\begin{table}
\centering
\caption{Dataset statistics.}\label{tab:app_dataset}
\begin{tabular}{lcrrrr}
    \toprule
    Dataset  & Vertices & Edges & Classes & Features\\
    \midrule
    Cora       &   2,708    &    5,429     &  7   & 1433 \\
    CiteSeer   &   3,327    &    4,732     &  6   & 3703 \\
    PubMed     &   19,717   &    44,338    &  3   & 500  \\
    Large Cora &   11,881   &    64,898    &  10  & 3780 \\
    NELL       &   65,755   &    266,144   & 210  & 5414 \\
    \bottomrule
\end{tabular}
\end{table}

\noindent \textbf{Baselines.}
We compare GLP and IGCN with the state-of-the-art semi-supervised classification methods: manifold regularization (ManiReg) \cite{belkin2006manifold}, semi-supervised embedding (SemiEmb) \cite{weston2012deep}, DeepWalk \cite{perozzi2014deepwalk}, iterative classification algorithm (ICA) \cite{sen2008collective}, Planetoid \cite{yang2016revisiting}, graph attention networks (GAT) \cite{velickovic2017graph}, multi-layer perceptron (MLP), LP \cite{Wu12parw}, and GCN \cite{kipf2016semi}. 

\noindent \textbf{Settings.} We use MLP as the classifier of GLP, and test GLP and IGCN with RNM and AR filters. We follow \cite{kipf2016semi} to use a two-layer structure for all neural networks, including MLP, GCN, IGCN. Guided by our analysis in \cref{sec:design}, the filter parameters $k$ and $\alpha$ should be set large if the label rate is small, and should be set small if the label rate is large.
Specifically, when 20 labels per class on citation networks are available or 10\% entities of NELL are labeled, we set $k=5$ for RNM and $\alpha=10$ for AR filters in GLP. In other scenarios with less labels, we set $k=10, \alpha=20$ for GLP. The $k, \alpha$ choosen for IGCN is equal to the above $k, \alpha$ divided by the number of layers -- 2.
We follow \cite{kipf2016semi} to set the parameters of MLP, GCN, IGCN: for citation networks, we use a two-layer network with 16 hidden units, $0.01$ learning rate, $0.5$ dropout rate, and $5\times10^{-4}$ L2 regularization, except that the hidden layer is enlarged to 64 units for Large Cora; for NELL, we use a two-layer network with 64 hidden units, $0.01$ learning rate, $0.1$ dropout rate, and $1\times10^{-5}$ L2 regularization. For more fair comparison with different baselines, we do not use a validation set for model selection as in \cite{kipf2016semi}, instead we select the model with the lowest training loss in 200 steps. All results are averaged over 50 random splits of the dataset. We set $\alpha$ of LP to $100$ for citation networks and $10$ for NELL. Parameters of GAT are same as  \cite{velickovic2017graph}. Results of other baselines are taken from \cite{yang2016revisiting,kipf2016semi}.

\begin{table*}[t]
    \centering
    \caption{Classification accuracy and running time on citation networks and NELL.}\label{tab:cite_nell}
    \small
    \setlength{\tabcolsep}{2pt}
    \begin{tabular}{l llll c llll c lll}
    \toprule
    {Label rate} & \multicolumn{4}{c}{{20 labels per class}} & & \multicolumn{4}{c}{{4 labels per class}} & & {10\%}& {1\%}& {0.1\%} \\
			\cmidrule(){2-5} \cmidrule(){7-10} \cmidrule(){12-14}
    \cmidrule(){2-5} \cmidrule(){7-10} \cmidrule(){12-14}
    & {Cora} & {CiteSeer} & {PubMed} & {Large Cora} & & {Cora} & {CiteSeer} & {PubMed} & {Large Cora} & & \multicolumn{3}{c}{{NELL}}\\
    \midrule
    {ManiReg}   & 59.5 & 60.1 & 70.7 & - & & - & - & - & - & & 63.4 & 41.3 & 21.8 \\
    {SemiEmb}   & 59.0 & 59.6 & 71.7 & - & & - & - & - & - & & 65.4 & 43.8 & 26.7 \\
    {DeepWalk}  & 67.2 & 43.2 & 65.3 & - & & - & - & - & - & & 79.5 & 72.5 & 58.1 \\
    {ICA}       & 75.1 & \textbf{69.1} & 73.9 & - & & 62.2 & 49.6 & 57.4 & - & & - & - & -\\
    {Planetoid} & 75.7 & 64.7 & 77.2 & - & & 43.2 & 47.8 & 64.0 & - & & 84.5 & 75.7 & 61.9 \\
    GAT & 79.5 & 68.2 & 76.2 & 67.4 &  & 66.6 & 55.0 & 64.6 & 46.4  & & - & - & -\\
    MLP & 55.1 ({\tiny0.6s}) & 55.4 ({\tiny0.6s}) & 69.5 ({\tiny0.6s}) & 48.0 ({\tiny0.8s}) & & 36.4 ({\tiny0.6s}) & 38.0 ({\tiny0.5s}) & 57.0 ({\tiny0.6s}) & 30.8 ({\tiny0.6s}) &  & 63.6 ({\tiny2.1s}) & 41.6 ({\tiny1.1s}) & 16.7 ({\tiny1.0s}) \\
    LP & 68.8 ({\tiny0.1s}) & 48.0 ({\tiny0.1s}) & 72.6 ({\tiny0.1s}) & 52.5 ({\tiny0.1s}) & & 56.6 ({\tiny0.1s}) & 39.5 ({\tiny0.1s}) & 61.0 ({\tiny0.1s}) & 37.0 ({\tiny0.1s}) &  & 84.5 ({\tiny0.7s}) & 75.1 ({\tiny1.8s}) & \textbf {65.9} ({\tiny1.9s}) \\
    GCN & 79.9 ({\tiny1.3s}) & 68.6 ({\tiny1.7s}) & \textbf {77.6} ({\tiny9.6s}) & 67.7 ({\tiny7.5s}) & & 65.2 ({\tiny1.3s}) & 55.5 ({\tiny1.7s}) & 67.7 ({\tiny9.8s}) & 48.3 ({\tiny7.4s}) &  & 81.6 ({\tiny33.5s}) & 63.9 ({\tiny33.5s}) & 40.7 ({\tiny33.2s}) \\
    \midrule
    IGCN({\scriptsize RNM}) & \textbf {80.9} ({\tiny1.2s}) & 69.0 ({\tiny1.7s}) & 77.3 ({\tiny10.0s}) & \textbf {68.9} ({\tiny7.9s}) & & \textbf {70.3} ({\tiny1.3s}) & \textbf {57.4} ({\tiny1.7s}) & \textbf {69.3} ({\tiny10.3s}) & \textbf {52.1} ({\tiny8.1s}) &  & \textbf {85.9} ({\tiny42.4s}) & \textbf {76.7} ({\tiny44.0s}) & \textbf {66.0} ({\tiny46.6s}) \\
    IGCN({\scriptsize AR}) & \textbf {81.1} ({\tiny2.2s}) & \textbf {69.3} ({\tiny2.6s}) & \textbf {78.2} ({\tiny11.9s}) & \textbf {69.2} ({\tiny11.0s}) & & \textbf {70.3} ({\tiny3.0s}) & \textbf {58.0} ({\tiny3.4s}) & \textbf {70.1} ({\tiny13.6s}) & \textbf {52.5} ({\tiny13.6s}) &  & \textbf {85.4} ({\tiny77.9s}) & \textbf {75.7} ({\tiny116.0s}) & \textbf {67.4} ({\tiny116.0s}) \\
    GLP({\scriptsize RNM}) & 80.3 ({\tiny0.9s}) & 68.8 ({\tiny1.0s}) & 77.1 ({\tiny0.6s}) & 68.4 ({\tiny1.8s}) & & \textbf {68.0} ({\tiny0.7s}) & 56.7 ({\tiny0.8s}) & 68.7 ({\tiny0.6s}) & 51.1 ({\tiny1.1s}) &  & \textbf {86.0} ({\tiny35.9s}) & \textbf {76.1} ({\tiny37.3s}) & 65.4 ({\tiny38.5s}) \\
    GLP({\scriptsize AR}) & \textbf {80.8} ({\tiny1.0s}) & \textbf {69.3} ({\tiny1.2s}) & \textbf {78.1} ({\tiny0.7s}) & \textbf {69.0} ({\tiny2.4s}) & & 67.5 ({\tiny0.8s}) & \textbf {57.3} ({\tiny1.1s}) & \textbf {69.7} ({\tiny0.8s}) & \textbf {51.6} ({\tiny2.3s}) &  & 80.3 ({\tiny57.4s}) & 67.4 ({\tiny76.6s}) & 55.2 ({\tiny78.6s}) \\
    \bottomrule
\end{tabular}
\end{table*}
\begin{table*}[ht]
    \centering
    \small
    \setlength{\tabcolsep}{2pt}
    \renewcommand{\arraystretch}{0.8}
    \caption{Results for unseen classes in AWA2.}
    \label{tab:awa2}
    \begin{tabular}{llllllllllclll}
    \toprule
    \multirow{2}{*}{Method} & \multirow{2}{*}{Devise} & \multirow{2}{*}{SYNC} & \multirow{2}{*}{GCNZ} & \multirow{2}{*}{GPM} & \multirow{2}{*}{DGPM} & \multirow{2}{*}{ADGPM} &  \multicolumn{3}{c}{IGCN({\scriptsize RNM})} & & \multicolumn{3}{c}{GLP({\scriptsize RNM})} \\
    \cmidrule(){8-10} \cmidrule(){12-14}
    & & & & & & & k=1 & k=2 & k=3 & & k=2 & k=4 & k=6 \\
\midrule
Accuracy & 59.7 & 46.6 &
68.0 ({\tiny1840s})  &
\textbf{77.3} ({\tiny864s})  &
67.2 ({\tiny932s})  &
76.0 ({\tiny3527s})  &
\textbf{77.9} ({\tiny864s})  &
\textbf{77.7} ({\tiny1583s})  &
73.1 ({\tiny2122s})  & &
76.0 ({\tiny12s})  &
75.0 ({\tiny13s})  &
73.0 ({\tiny11s}) \\
\bottomrule
    \end{tabular}
\end{table*}

\noindent \textbf{Performance of GLP and IGCN.}
The results are summarized in \cref{tab:cite_nell}, where the top 3 classification accuracies are highlighted in bold. Overall, GLP and IGCN perform best. Especially when the label rates are very small, they significantly outperform the baselines. Specifically, on citation networks, with 20 labels per class, GLP and IGCN perform slightly better than GCN and GAT, but outperform other baselines by a considerable margin. With 4 labels per class, GLP and IGCN significantly outperform all the baselines, demonstrating their label efficiency. On NELL, GLP and IGCN with the RNM filter as well as IGCN with the AR filter slightly outperform two very strong baselines -- LP and Planetoid, and outperform other baselines by a large margin.

\cref{tab:cite_nell} also reports the running time of the methods tested by us. We can see that GLP with the RNM filter runs much faster than GCN on most cases, and IGCN with the RNM filter has similar time efficiency as GCN.

\noindent \textbf{Results Analysis.}
Compared with methods that only use graph information, e.g., LP and DeepWalk, the large performance gains of GLP and IGCN clearly come from leveraging both graph and feature information. Compared with methods that use both graph and feature information, e.g., GCN and GAT, GLP and IGCN are much more label efficient. The reason is that they allow using stronger filters to extract higher level data representations to improve performance when label rates are low, which can be easily achieved by increasing the filter parameters $k$ and $\alpha$. But this cannot be easily achieved in the original GCN. As explained in \cref{sec:gcn}, to increase smoothness, GCN needs to stack many layers, but a deep GCN is difficult to train with few labels.

\subsection{Semi-Supervised Regression}
The proposed GLP and IGCN methods can also be used for semi-supervised regression. In \cite{wang2018zero}, GCN was used for zero-shot image recognition with a regression loss. Here, we replace the GCN model used in \cite{wang2018zero} with GLP and IGCN to test their performance on the zero-shot image recognition task.

Zero-shot image recognition in \cite{wang2018zero} is to learn a visual classifier for the categories with zero training examples, with only text descriptions of categories and relationships between categories. In particular, given a pre-trained CNN for known categories, \cite{wang2018zero} proposes to use a GCN to learn the model/classifier weights of unseen categories in the last-layer of the CNN. It first takes the word embedding of each category and the relations among all the categories (WordNet knowledge graph) as the inputs of GCN, then trains the GCN with the model weights of known categories in the last-layer of the CNN, and finally predicts the model weights of unseen categories.

\noindent \textbf{Datasets.}
We evaluate our methods and baselines on the ImageNet \cite{russakovsky2015imagenet} benchmark. ImageNet is an image database organized according to the WordNet hierarchy. All categories of ImageNet form a graph through ``is a kind of'' relation. For example, drawbridge is a kind of bridge, bridge is a kind of construction, and construction is a kind of artifact. According to \cite{wang2018zero}, the word embedding of each category is learned from Wikipedia by the GloVe text model \cite{pennington2014glove}.

\noindent \textbf{Baselines.}
We compare our methods GLP and IGCN with six state-of-the-art zero-shot image recognition methods, namely Devise \cite{frome2013devise}, SYNC \cite{changpinyo2016synthesized}, GCNZ \cite{wang2018zero}, GPM \cite{kampffmeyer2018rethinking}, DGPM \cite{kampffmeyer2018rethinking} and ADGPM \cite{kampffmeyer2018rethinking}. The prediction accuracy of these baselines are taken from their papers. Notably, the GPM model is exactly our IGCN with $k=1$.

\noindent \textbf{Settings.}
There are 21K different classes in ImageNet. We split them into a training set and a test set similarly as in \cite{kampffmeyer2018rethinking}. A ResNet-50 model was pre-trained on the ImageNet 2012 with 1k classes. The weights of these 1000 classes in the last layer of CNN are used to train GLP and IGCN for predicting the weights of the remaining classes. The evaluation of zero-shot image recognition is conducted on the AWA2 dataset \cite{xian2018zero}, which is a subset of ImageNet. For IGCN and the classifier (MLP) of GLP, we both use a two-layer structure with 2048 hidden units. We test $k=1,2,3$ for IGCN and $k=2,4,6$ for GLP. Results are averaged over 20 runs.

\noindent \textbf{Performance and Results Analysis.}
The results are summarized in \cref{tab:awa2}, where the top 3 classification accuracies are highlighted in bold. We can see that IGCN with $k=1,2$ and GPM \cite{kampffmeyer2018rethinking} perform the best, and outperform other baselines including Devise \cite{frome2013devise}, SYNC \cite{changpinyo2016synthesized}, GCNZ \cite{wang2018zero} and DGPM \cite{kampffmeyer2018rethinking} by a significant margin. GLP with $k=2$ is the second best compared with the baselines, only slightly lower than GPM. We observe that smaller $k$ achieves better performance on this task, which is probably because the diversity of features (classifier weights) should be preserved for the regression task \cite{kampffmeyer2018rethinking}. This also explains why DGPM \cite{kampffmeyer2018rethinking} (that expands the node neighborhood by adding distant nodes) does not perform very well. It is also worth noting that by replacing the 6-layer GCN in GCNZ with a 2-layer IGCN with $k=3$ and a GLP with $k=6$, the performance boosts from 68\% to around 73\%, demonstrating the low complexity and training efficiency of our methods. Another thing to notice is that GLP runs hundreds of times faster than GCNZ, and tens of times faster than others.

\section{Related Works}\label{sec:related}

There is a vast literature on semi-supervised learning \cite{chapelle2006semi,zhu2009introduction}, including generative models \cite{DBLP:conf/nips/Baluja98a,kingma2014semi}, semi-supervised support vector machine \cite{DBLP:conf/nips/BennettD98}, self-training \cite{DBLP:conf/uai/HaffariS07}, co-training \cite{DBLP:conf/colt/BlumM98}, and graph-based methods \cite{DBLP:conf/nips/KapoorQAP05,LiWZ16,LiangL18,ZhangZYL16}.

Early graph-based methods adopt a common assumption that nearby vertices are likely to have same labels. One approach is to learn smooth low-dimensional embeddings with Markov random walks \cite{szummer2002partially}, Laplacian eigenmaps \cite{belkin2004semi}, spectral kernels \cite{chapelle03,zhang06}, and context-based methods \cite{perozzi2014deepwalk}. Another line of works rely on graph partition, where the cuts should agree with the labeled vertices and be placed in low density regions \cite{blum2001learning,blum2004semi,joachims2003transductive,Zhu03}, among which the most popular one is perhaps the label propagation methods \cite{bengio2006label,chapelle2005semi,Zhou03}. It was shown in \cite{ekambaram2013wavelet,girault2014semi} that they can be interpreted as low-pass graph filtering. To further improve learning performance, many methods were proposed to jointly model graph structures and data features. Iterative classification algorithm \cite{sen2008collective} iteratively classifies an unlabeled data by using its neighbors' labels and features. Manifold regularization \cite{belkin2006manifold}, deep semi-supervised embedding \cite{weston2012deep}, and Planetoid \cite{yang2016revisiting} regularize a supervised classifier with a Laplacian regularizer or an embedding-based regularizer.

Inspired by the success of convolutional neural networks (CNN) on grid-structured data such as image and video, a series of works proposed a variety of graph convolutional neural networks \cite{bruna2014spectral,henaff2015deep,duvenaud2015convolutional,atwood2016diffusion} to extend CNN to general graph-structured data. To avoid the expensive eigen-decomposition, ChebyNet \cite{defferrard2016convolutional} uses a polynomial filter represented by $k$-th order polynomials of graph Laplacian via Chebyshev expansion. Graph convolutional networks (GCN) \cite{kipf2016semi} further simplifies ChebyNet by using a localized first-order approximation of spectral graph convolution, and has achieved promising results in semi-supervised learning. It was shown in \cite{li2018deeper} that the success of GCN is due to performing Laplacian smoothing on data features. MoNet \cite{MontiBMRSB17} shows that various non-Euclidean CNN methods including GCN are its particular instances. Other related works include GraphSAGE \cite{hamilton2017inductive}, graph attention networks \cite{velickovic2017graph}, attention-based graph neural network \cite{thekumparampil2018attention}, graph partition neural networks \cite{liao2018graph}, FastGCN \cite{ChenMX18}, dual graph convolutional neural network  \cite{zhuang2018dual}, stochastic GCN \cite{ChenZS18}, Bayesian GCN \cite{zhang2018bayesian}, deep graph infomax \cite{velivckovic2018deep}, LanczosNet \cite{liao2019lanczosnet}, etc.  We refer readers to two comprehensive surveys \cite{corr/abs-1812-04202,corr/abs-1812-08434} for more discussions.

Another related line of research is feature smoothing, which has long been used in computer graphics for fairing 3D surface \cite{taubin1995curve}. \cite{Hein07} proposed manifold denoising (MD) by using feature smoothing as a preprocessing step for semi-supervised learning, where the denoised data features are used to construct a graph for running a label propagation algorithm. MD uses the data features to construct a graph and employs the AR filter for feature smoothing. However, it cannot be directly applied to datasets such as citation networks where the graph is given.

\section{Conclusion}\label{sec:conclusion}

This paper studies semi-supervised learning from a unifying graph filtering perspective, which offers new insights into the classical label propagation methods and the recently popular graph convolution networks. Based on the analysis, we propose generalized label propagation methods and improved graph convolutional networks to extend their modeling capabilities and achieve label efficiency. In the future, we plan to investigate the design and automatic selection of proper graph filters for various application scenarios and apply the proposed methods to solve more real applications.

\section*{Acknowledgments}

This research was supported by the grants 1-ZVJJ and G-YBXV funded by the Hong Kong Polytechnic University.


{\small
\bibliographystyle{ieee}
\bibliography{parw}
}
\end{document}